\documentclass[10pt,twocolumn,letterpaper]{article}

\usepackage{cvpr}              
\usepackage[accsupp]{axessibility}

\definecolor{cvprblue}{rgb}{0.21,0.49,0.74}
\usepackage[pagebackref,breaklinks,colorlinks,allcolors=cvprblue]{hyperref}

\def\paperID{26} 
\def\confName{CVPR}
\def\confYear{2026}

\title{Bridging the Training-Deployment Gap: Gated Encoding and Multi-Scale Refinement for Efficient Quantization-Aware Image Enhancement}

\author{
Dat To-Thanh$^{1, 5}$ \quad Nghia Nguyen-Trong$^{2, 5}$ \quad Hoang Vo$^{1,5  }$ \quad Hieu Bui-Minh$^{3}$ \quad Tinh-Anh Nguyen-Nhu$^{4, 5\dagger}$ \\ \\
$^{1}$ University of Science, VNU-HCM, Vietnam \\
$^{2}$ University of Information Technology, VNU-HCM, Vietnam \\
$^{3}$ Da Nang University of Economics, Vietnam \\
$^{4}$ Ho Chi Minh University of Technology, VNU-HCM, Vietnam \\
$^{5}$ Vietnam National University, Ho Chi Minh City, Vietnam \\
$^{\dagger}$ Corresponding author\\
}

\begin{document}

\maketitle
\begin{abstract}
Image enhancement models for mobile devices often struggle to balance high output quality with the fast processing speeds required by mobile hardware. While recent deep learning models can enhance low-quality mobile photos into high-quality images, their performance is often degraded when converted to lower-precision formats for actual use on mobile phones. To address this training-deployment mismatch, we propose an efficient image enhancement model designed specifically for mobile deployment. Our approach uses a hierarchical network architecture with gated encoder blocks and multiscale refinement to preserve fine-grained visual features. Moreover, we incorporate Quantization-Aware Training (QAT) to simulate the effects of low-precision representation during the training process. This allows the network to adapt and prevents the typical drop in quality seen with standard post-training quantization (PTQ). Experimental results demonstrate that the proposed method produces high-fidelity visual output while maintaining the low computational overhead needed for practical use on standard mobile devices. The code will be available
at \href{https://github.com/GenAI4E/QATIE.git}{https://github.com/GenAI4E/QATIE.git}.
\end{abstract}    
\section{Introduction}
\label{sec:intro}

Recent advances in deep learning and computer vision have led to remarkable performance across a wide range of tasks \cite{vu2025modelvisualquestionanswering,nguyen2025seehearunderstandbenchmarking}, including image retrieval \cite{nguyen2026itselfattentionguidedfinegrained,nguyen2025hybridunifiediterativenovel}, visual question answering \cite{nguyennhu2025stervlmspatiotemporalenhancedreference,nguyen2024improvinggeneralizationvisualreasoning}. Besides, smartphone photography has become the dominant imaging modality for billions of users, yet a substantial perceptual gap remains between images captured by low-end mobile devices and those produced by Digital Single-Lens Reflex (DSLR) cameras. This gap is rooted in hardware limitations, including small sensors and compact optics, which collectively degrade signal-to-noise ratio, dynamic range, color fidelity, and sharpness. To address these limitations, recent advances in deep learning have introduced \emph{Deep Image Signal Processing (Deep ISP)} models that learn end-to-end mappings from low-quality mobile images to high-quality DSLR outputs, effectively replacing or augmenting traditional ISP pipelines through data-driven optimization~\cite{ignatov2017dslr,schwartz2018deepisp}. 

In this work, we consider the image enhancement setting, where paired images are synchronously captured using an iPhone 3GS and a Canon 70D DSLR, enabling supervised learning of a direct smartphone-to-DSLR transformation~\cite{ignatov2017dslr}. This task requires the joint correction of noise, blur, color distortions, and tone inconsistencies in a single forward pass. Prior works~\cite{ignatov2017dslr,schwartz2018deepisp,chen2018learning,johnson2016perceptual} have demonstrated that combining pixel-wise, perceptual, and adversarial losses is critical for achieving visually plausible results in this setting. However, these approaches are typically developed and evaluated under full-precision assumptions, and existing quantization techniques do not adequately account for the distributional characteristics of Deep ISP features under real-world mobile deployment~\cite{jacob2018quantization,krishnamoorthi2018quantization}.

Some challenge reports~\cite{ignatov2018pirm, ignatov2025rgb} explicitly emphasize the strict runtime constraints on mobile hardware. In particular, models are required to process FullHD images on-device using mobile inference frameworks such as TensorFlow Lite~\cite{jacob2018quantization}, while maintaining high perceptual quality and avoiding visible artifacts. This introduces a fundamental \emph{training–deployment mismatch}: models are optimized in continuous full/half-precision domains (FP32/FP16), but executed under 8-bit quantized settings (i.e., INT8), where numerical precision and dynamic range are severely constrained. As a result, models that perform well during training often degrade significantly after deployment, especially when evaluated under realistic latency constraints.

This mismatch is particularly critical for high-fidelity pixel-level tasks such as image enhancement and super-resolution~\cite{lim2017edsr}. Unlike classification tasks, where feature representations are relatively robust to quantization noise, Deep ISP models exhibit highly sensitive activation distributions that are long-tailed, asymmetric, and strongly input-dependent~\cite{tu2023ptqsr,nagel2020datafree}. Moreover, activation outliers are often closely tied to color and luminance information, such that improper quantization can lead to perceptually severe artifacts, including color shifts, banding, and texture distortions. These properties suggest that existing quantization pipelines, when applied after full-precision training, are insufficient to maintain perceptual fidelity in Deep ISP. Moreover, existing QAT approaches are primarily designed for high-level vision tasks and do not explicitly model the color-sensitive and heavy-tailed activation distributions that are critical for pixel-level image reconstruction.

To address this challenge, we advocate a deployment-consistent optimization paradigm: \emph{optimize the model that will be deployed}. Specifically, we introduce Quantization-Aware Training (QAT) into the training framework, as shown in~\Cref{fig:qat}, allowing the model to learn representations that are inherently robust to quantization effects. Although QAT has been widely studied in high-level vision tasks, its application to Deep ISP has been largely unexplored in previous work~\cite{ignatov2017dslr, hui2018perception, schwartz2018deepisp}. By simulating quantization during training via fake quantization operators and gradient approximations~\cite{jacob2018quantization, krishnamoorthi2018quantization, bengio2013estimating}, QAT aligns the training objective with the deployment environment, mitigating the degradation introduced by 8-bit inference.

Our contributions are summarized as follows:

\begin{itemize}
    \item We identify the specific challenges of model quantization in pixel-level image enhancement, highlighting the causes of performance degradation during mobile deployment.
    \item We introduce a training framework that incorporates Quantization-Aware Training (QAT) specifically designed to minimize this degradation and ensure deployment-consistent optimization.
    \item We demonstrate the high scalability of our proposed models, showing that they maintain an effective balance between computational efficiency and perceptual output quality.
    \item We validate that our model exhibits low computational overhead and efficient resource management, ensuring feasibility for deployment on standard commercial mobile devices. 
\end{itemize}
\section{Related Works}
\label{sec:related_works}

\subsection{Deep ISP and Photo Enhancement}

Recent advances in deep learning have enabled \emph{Deep Image Signal Processing (Deep ISP)}, where end-to-end models directly learn mappings from raw or degraded inputs to high-quality sRGB outputs, effectively replacing traditional hand-crafted ISP pipelines \cite{heide2014flexisp,ignatov2017dslr,brooks2019unprocessing,phannguyen2025cycletrainingsemisuperviseddomain}. The DPED benchmark is a seminal work in this direction, introducing paired smartphone–DSLR data and demonstrating the importance of perceptual loss formulations for realistic enhancement \cite{ignatov2017dslr}. Subsequent work, such as DeepISP, further explores end-to-end ISP learning from RAW inputs \cite{schwartz2018deepisp}, while Learning to See in the Dark highlights the effectiveness of deep models under extreme low-light conditions \cite{chen2018learning}.

In parallel, image enhancement methods have explored diverse formulations to improve perceptual quality and robustness. Zero-DCE formulates enhancement as pixel-wise curve estimation without paired supervision \cite{guo2020zerodce}, while MIRNet leverages multi-scale feature fusion to capture complex spatial dependencies for restoration \cite{zamir2020mirnet}. Weakly supervised approaches such as WESPE relax alignment requirements but often introduce trade-offs in pixel-level fidelity and color consistency \cite{ignatov2017wespe}.

However, these methods are predominantly developed and evaluated under full-precision settings, and do not explicitly consider deployment constraints or robustness under quantized inference, limiting their applicability in real-world mobile scenarios.

\subsection{Efficient Mobile Vision and Quantization}

Quantization has emerged as a key technique for efficient inference on edge devices~\cite{choi2018pact, esser2019lsq, jacob2018quantization, krishnamoorthi2018quantization, nah2017deep}. Integer-only inference frameworks demonstrate that lower-bit representations can significantly improve latency and energy efficiency when co-designed with hardware \cite{jacob2018quantization}. Quantization-Aware Training (QAT) further improves accuracy by incorporating quantization effects during training, with representative methods such as LSQ learning quantization step sizes \cite{esser2019lsq} and PACT optimizing activation clipping \cite{choi2018pact}.

Despite these advances, existing quantization methods are primarily designed for high-level vision tasks such as classification, where global semantic representations are relatively robust to quantization noise \cite{krishnamoorthi2018quantization}. They do not explicitly account for the sensitivity of pixel-level reconstruction tasks, where small perturbations can lead to noticeable visual degradation \cite{nah2017deep}.

\subsection{Quantization for Image Restoration}

Quantizing image restoration models is highly challenging because their activation distributions are non-Gaussian, long-tailed, asymmetric, and strongly input-dependent~\cite{tu2023ptqsr, liu20242dquant}. To address this in super-resolution, recent post-training quantization (PTQ) approaches rely on targeted calibration, such as using density-based dual clipping with pixel-aware calibration~\cite{tu2023ptqsr} or dual-stage bound initialization refined by knowledge distillation~\cite{liu20242dquant, hinton2015distilling}. While effective for super-resolution, these joint optimization strategies remain largely underexplored for Deep ISP and high-fidelity photo enhancement. Because extreme activations in these models are closely tied to color representation, standard quantization can introduce severe visual artifacts, introducing a critical need for new techniques that explicitly combine Deep ISP architectures with quantization-aware optimization.

\begin{figure*}[htbp]
    \centering
    \includegraphics[width=0.9\textwidth]{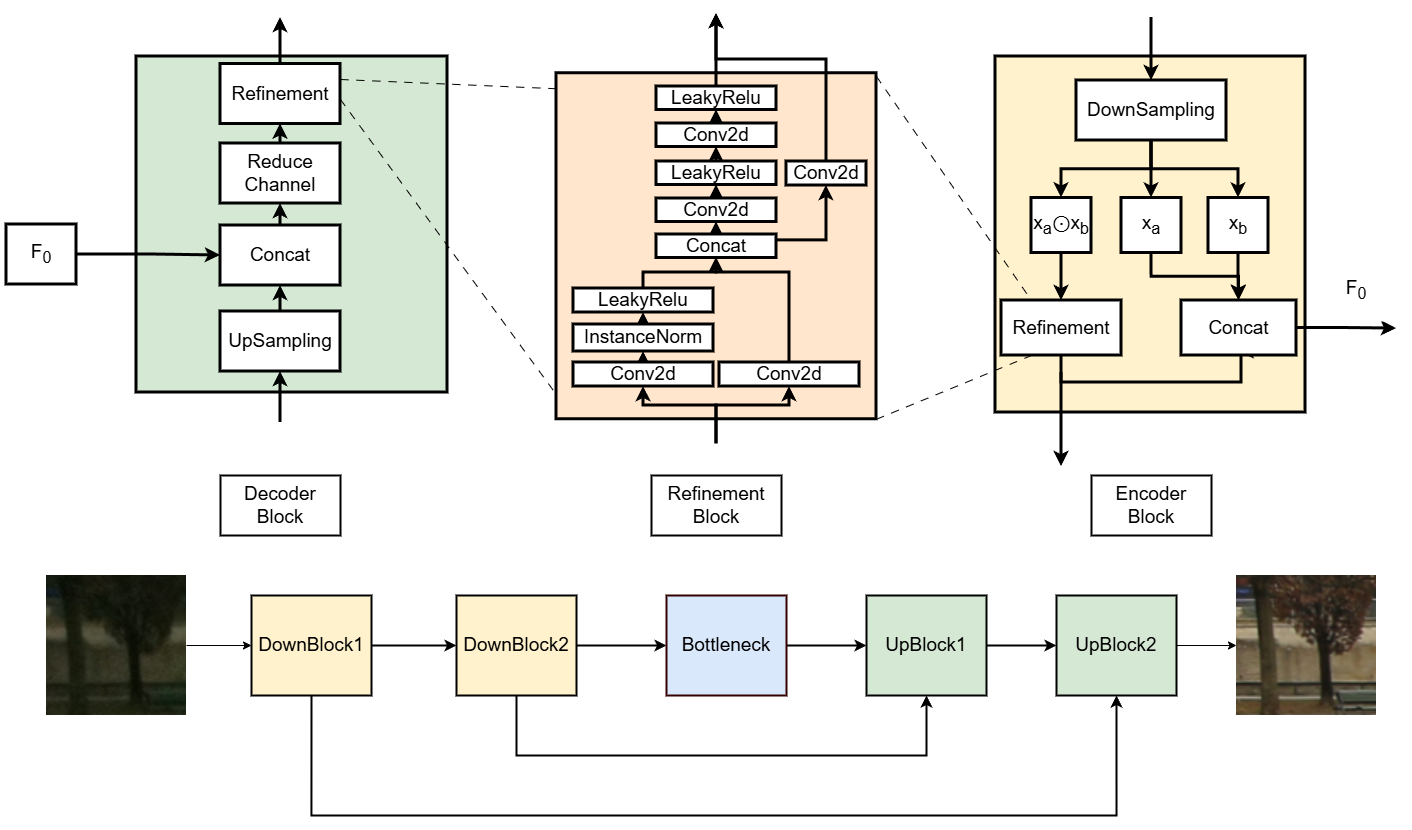}
    \caption{The proposed hybrid architecture for RGB image enhancement. The model features a three-scale hierarchical structure comprising Gated Encoder Blocks with multichannel skip connections, Multi-Scale Refinement modules applied at scales $S/2$, $S/4$, and $S/8$, and a Multi-branch Feature Fusion strategy in the decoder to aggregate multi-level semantic features for high-fidelity restoration.}
    \label{fig:pipeline}
\end{figure*}

\section{Proposed Method}

In this section, we detail the proposed hybrid architecture for RGB image enhancement, the choice of loss functions and the Quantization-Aware Training (QAT) strategies used to achieve high-fidelity enhancement. An overview of the proposed framework is illustrated in \textbf{\Cref{fig:pipeline}}.

\subsection{Network Architecture}
The proposed model adopts a three-scale hierarchical structure to effectively balance global context aggregation and fine-grained texture preservation. 

\subsubsection{Gated Encoder Block}

The encoder block, inspired by the Downblock architecture provided by the DaHua-IIG team \cite{ignatov2025rgb}, is designed to progressively reduce spatial resolution while capturing rich semantic hierarchies. Each down-sampling stage is implemented via a \textit{Gated Down-sampling Block}, which employs a dual-branch architecture to perform simultaneous feature extraction and non-linear gating. In our specific architecture, these blocks are sequentially applied to compress the original spatial resolution $S$ down to intermediate $S/2$ and $S/4$ scales.

Specifically, given an input tensor $\mathbf{X} \in \mathbb{R}^{C_{in} \times H \times W}$, the block utilizes two parallel convolutional branches, $\mathcal{F}_a$ and $\mathcal{F}_b$. The intermediate feature maps, $\mathbf{x}_a$ and $\mathbf{x}_b$, are obtained by applying a hyperbolic tangent activation function:

\begin{equation}
    \mathbf{x}_a = \tanh(\mathcal{F}_a(\mathbf{X})), \quad \mathbf{x}_b = \tanh(\mathcal{F}_b(\mathbf{X}))
\end{equation}

To effectively modulate the information flow, we introduce an element-wise gating mechanism. The primary gated feature stream, $\mathbf{x}_g$, is calculated as the Hadamard product of the two branches:

\begin{equation}
    \mathbf{x}_g = \mathbf{x}_a \odot \mathbf{x}_b
\end{equation}

This gating operation serves as a lightweight spatial-channel attention mechanism, where $\mathbf{x}_b$ can be interpreted as a soft mask that filters the activations of $\mathbf{x}_a$ (and vice versa).

Unlike standard architectures that only propagate the final output of a block, our Gated Encoder preserves the triplet $(\mathbf{x}_a, \mathbf{x}_g, \mathbf{x}_b)$ at each stage. This entire triplet is forwarded to the corresponding decoder layer via a \textbf{multichannel skip connection}. While $\mathbf{x}_g$ continues through the main encoding path to form deeper high-level representations, the inclusion of $\mathbf{x}_a$ and $\mathbf{x}_b$ in the skip connection ensures that the decoder retains access to both filtered semantic information and raw directional features. This multi-stream redundancy facilitates a more precise reconstruction of fine-grained spatial details that might otherwise be suppressed by the gating operation.

\subsubsection{Multi-Scale Refinement}

To improve feature quality across different spatial resolutions, we leverage a \textbf{Multi-Scale Refinement} strategy \cite{10473166} that is consistently applied throughout both the encoder and decoder pathways. This design enables the network to jointly preserve the global illumination structure and local texture details, which is particularly important for image enhancement.

Formally, after each gated down-sampling stage, the main propagated feature $\mathbf{x}_g$ is refined as
\begin{equation}
    \tilde{\mathbf{F}} = \mathcal{R}(\mathbf{x}_g),
\end{equation}
where $\mathcal{R}(\cdot)$ denotes the refinement operator implemented by a UNet-style residual convolutional block. In the proposed model, this refinement is applied at scale $S/2$ and $S/4$ in the encoder, as well as at the bottleneck representation at scale $S/8$. As a result, the network progressively enhances feature discriminability while preserving structurally meaningful information across scales.

Each refinement block is composed of a sequence of convolutional transformations with residual learning. Given an input feature $\mathbf{F}$, the block first applies a $3\times3$ convolution followed by instance normalization and LeakyReLU activation to generate an intermediate representation. In parallel, a lightweight $1\times1$ convolution is applied to the original feature to preserve low-level information. The two streams are concatenated and further processed by stacked convolutions, while an additional $1\times1$ projection is used to form a residual shortcut. The final refined feature is obtained by adding the transformed feature and the shortcut output. This design improves representation capacity without significantly increasing computational cost.

In the decoder, multi-scale refinement is performed after feature fusion at each reconstruction stage. Specifically, the upsampled decoder feature, the refined encoder feature at the corresponding scale, and the preserved skip feature are concatenated and first compressed by a $1\times1$ convolution. The reduced feature is then passed through another refinement block:
\begin{equation}
    \hat{\mathbf{F}} = \mathcal{R}\big(\phi([\mathbf{F}_{up}, \mathbf{F}_{enc}, \mathbf{F}_{skip}])\big),
\end{equation}
where $[\cdot]$ denotes channel-wise concatenation and $\phi(\cdot)$ is the channel reduction operator. This refinement after fusion helps to reconcile coming information from different semantic levels, leading to a more stable reconstruction and better preservation of image details.

By inserting refinement modules at multiple resolutions - namely $S/2$, $S/4$, and the bottleneck scale $S/8$ - the model can simultaneously capture coarse illumination patterns and fine-grained local structures. This hierarchical refinement mechanism is crucial for enhancing images, where both global brightness adjustment and local detail recovery must be handled in a unified manner.

\subsubsection{Decoder and Feature Fusion}

The decoder path \cite{ignatov2025rgb} is designed to restore spatial resolution and reconstruct the enhanced image by aggregating features from multiple semantic levels. This recovery process is facilitated through successive up-sampling stages combined with a specialized \textbf{Multibranch Feature Fusion} strategy. At each decoding scale, the network ensures a comprehensive and feature-rich reconstruction by concatenating three distinct information streams along the channel dimension. Specifically, the fusion integrates high-level semantic context propagated as \textbf{upsampled features} from the deeper decoder layers, \textbf{refined encoder features} that have been processed through a UNet-style convolutional block to maintain semantic consistency, and \textbf{raw skip features} consisting of the concatenated directional cues $(\mathbf{x}_a, \mathbf{x}_b)$ from the encoder's gated branches. This triple-stream concatenation allows the model to recover fine-grained spatial details while maintaining global structural integrity.

Because this concatenation significantly expands the channel capacity, a $1 \times 1$ convolutional layer is immediately applied. This pointwise convolution acts as a channel-wise dimensionality reduction, compressing the channels back to a manageable size while allowing the network to learn optimal linear combinations of the three distinct feature streams.

Following this reduction, the fused tensor is passed through a Refinement Block. This dense convolutional stage harmonizes the aggregated features, smoothing out inconsistencies and ensuring structural integrity before the next up-sampling step.

\subsection{Loss Function}

To supervise the network across multiple dimensions, including pixel fidelity, structural consistency, and robustness, we employ Peak Signal-to-Noise Ratio (PSNR) loss, the Cosine Similarity loss, and the Outlier-Aware loss~\cite{yan2025mobileie}.

Given a ground truth image $I^{H \times W}$ and an output image from the model $\hat{I}^{H \times W}$, the normalized PSNR loss is formulated by first determining the Root Mean Squared Error (RMSE):

\begin{equation}
\text{RMSE} = \sqrt{\frac{1}{HW} \sum_{i=1}^{H} \sum_{j=1}^{W} (\hat{I}_{i,j} - I_{i,j})^2}
\end{equation}

Using the RMSE, the Peak Signal-to-Noise Ratio (PSNR) is calculated as follows:

\begin{equation} \label{eq:psnr}
\text{PSNR} = 20 \cdot \log_{10} \left( \frac{MAX_I}{\text{RMSE}} \right) 
\end{equation}

where $MAX_I$ represents the maximum possible pixel value (e.g., 1.0 or 255). Finally, to maximize the PSNR, the final loss function $\mathcal{L}_{PSNR}$ is formulated by normalizing the value to a $[0, 1]$ range (assuming a baseline of 50.0):

\begin{equation}
\mathcal{L}_{PSNR} = \frac{50.0 - \text{PSNR}}{100.0}
\end{equation}

The \textit{Cosine Similarity loss} enforces structural integrity by aligning the direction of outputs and ground truth images:
\begin{equation}
\mathcal{L}_{cos} = 1 - \frac{\mathbf{I} \cdot \mathbf{\hat{I}}}{\|\mathbf{I}\| \|\mathbf{\hat{I}}\|} = 1 - \frac{\sum_{i=1}^{n} I_i \hat{I}_i}{\sqrt{\sum_{i=1}^{n} I_i^2} \sqrt{\sum_{i=1}^{n} \hat{I}_i^2}}.
\end{equation}

The total loss function $\mathcal{L}_{total}$ is then formulated as:
\begin{equation} \label{eq:totalloss}
\mathcal{L}_{total} = \alpha \cdot \mathcal{L}_{PSNR} + \beta \cdot \mathcal{L}_{cos} + \gamma \cdot \mathcal{L}_{out},
\end{equation}
where $\mathcal{L}_{PSNR}$, $\mathcal{L}_{cos}$, and $\mathcal{L}_{out}$ represent the PSNR loss, the Cosine Similarity loss, and the Outlier-Aware loss, respectively. 

This joint objective ensures that the model recovers both global illumination and fine-grained textures while remaining robust against localized artifacts during the training phase. Specifically, the PSNR loss focuses on high-quality pixel reconstruction, the Cosine Similarity loss enforces directional alignment of feature vectors for structural integrity, and the Outlier-Aware loss is employed to stabilize the learning process by dynamically weighting pixels based on the error distribution, preventing the gradients from being dominated by extreme values.

\subsection{Quantization-Aware Training Optimization (QAT)}

\begin{figure*}[htbp]
    \centering
    \includegraphics[width=0.9\linewidth]{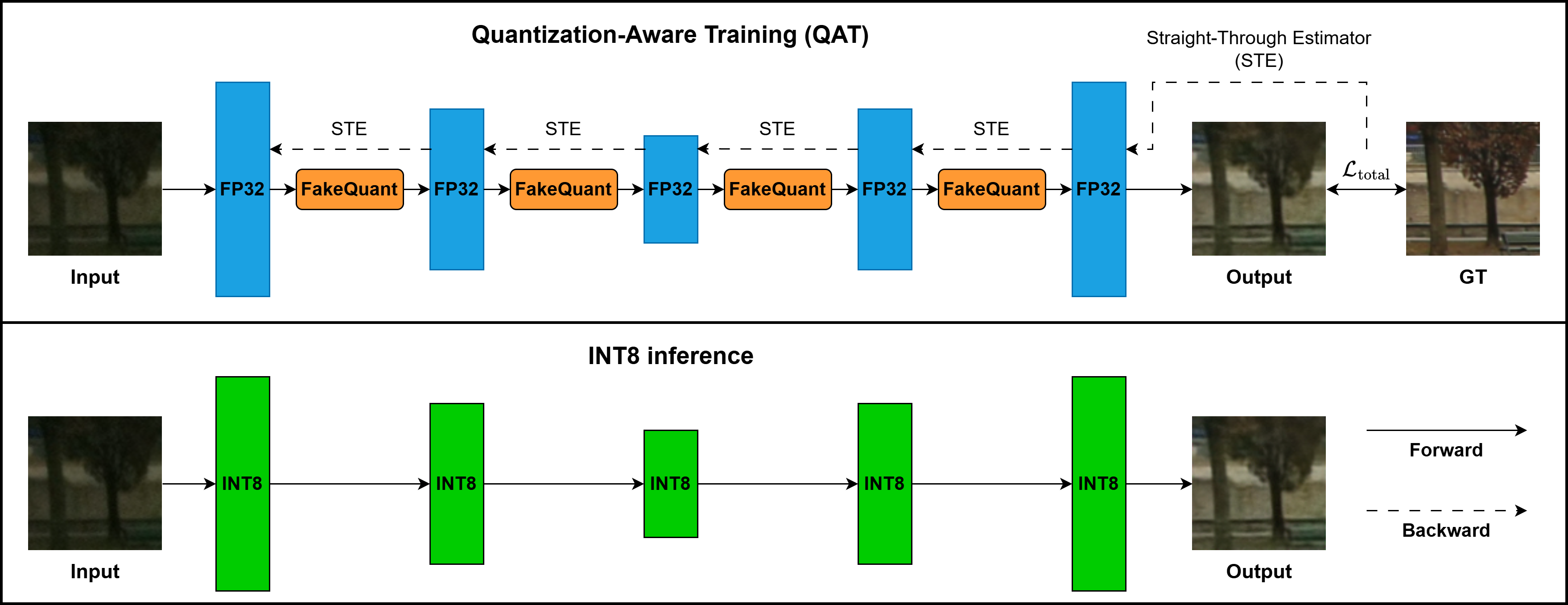}
    \caption{Quantization-Aware Training (QAT) and inference pipelines. (Top) QAT simulates low-precision effects via FakeQuant nodes during the forward pass while maintaining FP32 compute. To bypass non-differentiable quantization steps, the backward pass uses the Straight-Through Estimator (STE) to route gradients and update FP32 weights, minimizing the loss ($\mathcal{L}_{\text{total}}$) against the ground truth (GT). (Bottom) Post-training, the model is converted to a purely INT8 representation for efficient integer-only inference.}
    \label{fig:qat}
\end{figure*}

To ensure the proposed model is hardware-friendly and efficient for mobile deployment, we incorporate Quantization-Aware Training (QAT) as a final optimization stage, as illustrated in \textbf{\Cref{fig:qat}}. The core of this process is Fake Quantization (FakeQuant), which introduces simulated lower-precision operations, e.g. clamping and rounding, into the computational graph. Mechanically, as shown in the top pipeline of \textbf{\Cref{fig:qat}}, FakeQuant temporarily maps weights and activations to discrete levels during the forward pass while maintaining overall FP32 computation. Because discrete rounding is non-differentiable, the backward pass utilizes a Straight-Through Estimator (STE)~\cite{bengio2013estimatingpropagatinggradientsstochastic} to route gradients past these nodes, enabling continuous updates to the underlying FP32 weights. By simulating these hardware quantization effects during training, the network proactively learns to compensate for precision loss and rounding errors. This optimization phase refines the entire architecture using a reduced learning rate of 0.00001. Following QAT, the model is seamlessly converted into a INT8 representation (bottom of \textbf{\Cref{fig:qat}}), drastically reducing the memory footprint for mobile inference while preserving high-fidelity enhancement results.

\section{Experiments}
\label{sec:experiments}

\subsection{Experimental Settings}

\textbf{Implementation Details.} We implement our model in PyTorch and use the Pytorch Lightning library to train. We use the Adam optimizer with a cosine annealing learning rate schedule with warmup, starting at 0.00001. A 5-epoch warm-up phase is applied with learning rate gradually increases to 0.0001. The model is trained for more than 62k iterations (50 epochs) with an effective batch size is 128 and gradient accumulation of 2 steps. Furthermore, the model is trained with bfloat16 precision to reduce the training time and trained with gradient clipping to the [-1.0, 1.0] range for better training stability. We choose $\alpha=2.0, \beta=1.0, \gamma=1.0$ for weights of total loss \Cref{eq:totalloss}. All training are done on a single NVIDIA RTX A6000 GPU. Inference latency is evaluated by exporting models to the TFLite format and benchmarking them on a Snapdragon 8 Gen 2 smartphone with Full HD ($1920 \times 1080$) input images. All latencies are recorded using the AI Benchmark application~\cite{ignatov2019ai, 9022101}.

\textbf{Dataset and Metrics.} The model is trained primarily on more than 160k pairs of 100x100 patches taken from iPhone and Canon DSLR images from DPED~\cite{ignatov2017dslr} dataset. The evaluation is performed on a validation set consisting of 9310 pairs of 100x100 test patches taken from iPhone, Sony, Blackberry from the same dataset. We adopt two objective evaluation metrics: Peak Signal-to-Noise Ratio (PSNR) (using \textbf{\Cref{eq:psnr}}) and Structural SIMilarity (SSIM)~\cite{SSIMpaper}.

\subsection{Results}

\subsubsection{Quantitative Results}
As detailed in \textbf{\Cref{tab:leaderboard}}, our model achieved second place overall in the Mobile AI 2026 RGB Image Enhancement Challenge. With a PSNR of 21.82 dB and an SSIM of 0.7653, the model ensures high structural fidelity, while a MOS of 3.2 contributes to a competitive Final Score of 3.8. Consequently, our entry ranked second in the competition for both reconstruction precision and total score.

\begin{table*}[t]
  \centering
  \small
  \setlength{\tabcolsep}{3pt}
  \caption{Results of our proposed model and other methods on Mobile AI 2026 sRGB image enhancement challenge. The runtime values were obtained on 1024×1024 images.}
  \label{tab:leaderboard}
  \begin{tabular}{l|ccc|cc|c}
    \toprule
    Team  & PSNR$\uparrow$ & SSIM$\uparrow$ & MOS$\uparrow$ & Adreno GPU, ms$\downarrow$ & Arm GPU, ms$\downarrow$ & Final Score \\
    \midrule
    DaHua-IIG  & \textbf{22.20} & \textbf{0.7881} & \textbf{4.1} & \textbf{23.8} & \textbf{60.4} & \textbf{163.0} \\
    \textbf{Capybara (Our)} &  \underline{21.82} & \underline{0.7653} & \underline{3.2} & 291.0 & 266.0 & \underline{3.8} \\
    DH-XHDL-Team & 20.55 & 0.7601 & 1.2 & \underline{30.8} & \underline{52.4} & 0.28 \\
    \bottomrule
  \end{tabular}
\end{table*}

\subsubsection{Qualitative Results}

\begin{figure}
    \centering
    \includegraphics[width=0.5\textwidth]{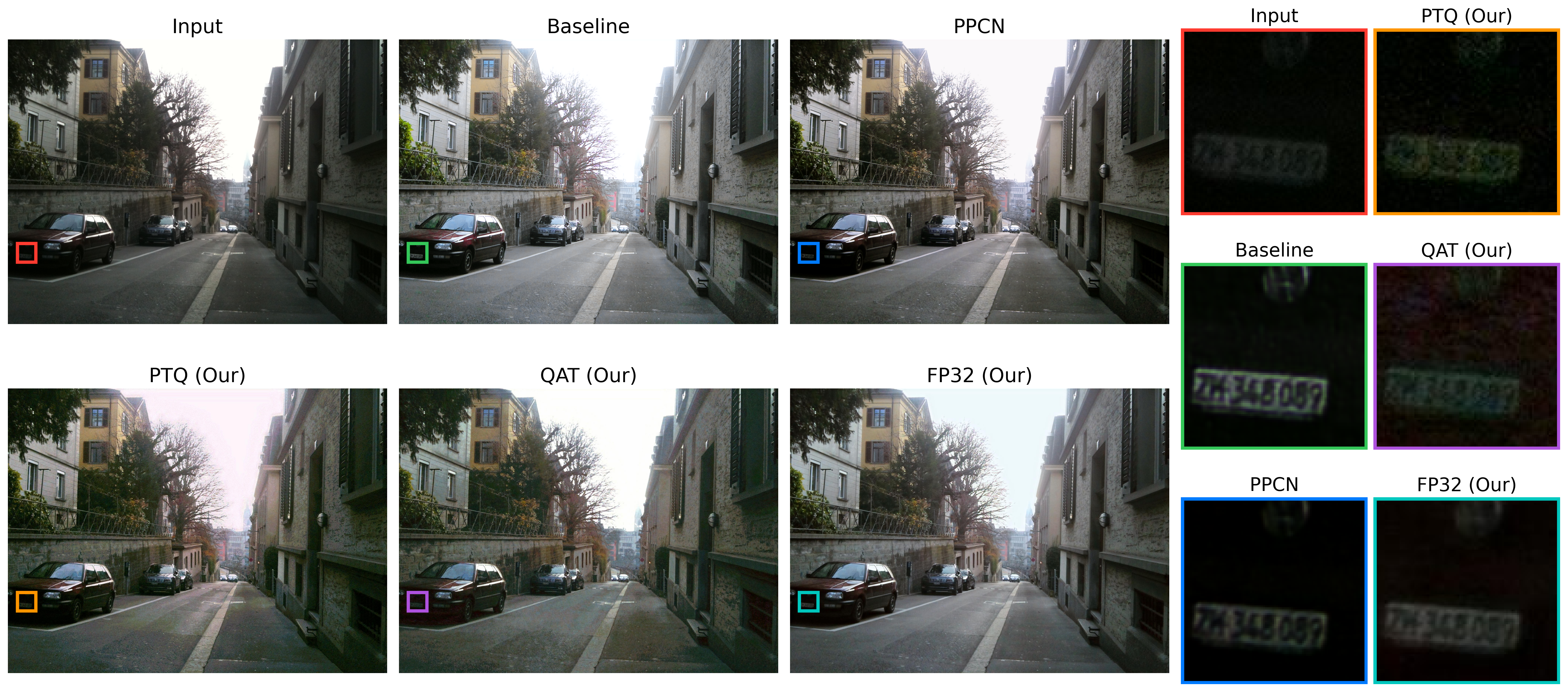}
    \caption{
    Qualitative results on image 01 of the full-size test subset of iPhone in the DPED dataset. The comparison includes the input image, baseline~\cite{ignatov2017dslr}, PPCN~\cite{hui2018perception}, and our models (direct 8-bit PTQ, 8-bit QAT, and FP32).
    }
    \label{fig:qualitative}
\end{figure}

\textbf{\Cref{fig:qualitative}} presents a qualitative comparison between the input image, the baseline~\cite{ignatov2017dslr}, PPCN~\cite{hui2018perception}, and our model evaluated on the DPED dataset. As observed in the zoomed-in crops, our full-precision (FP32) model effectively reconstructs sharp details and textures, such as the text on the license plate. This strong reconstruction capability allows it to achieve a visual quality highly comparable to the established baseline and PPCN methods. Furthermore, the comparison demonstrates the significant degradation that occurs when converting to 8-bit precision using a direct Post-Training Quantization (PTQ) model. This PTQ approach suffers from severe color shifts, such as the pink sky, alongside noticeable noise artifacts shown in the cropped part. In contrast, our 8-bit model trained with Quantization-Aware Training (QAT) successfully mitigates these issues, restoring overall color fidelity and fine structural details to recover a visual quality that closely matches the FP32 model.

\subsection{Ablation Studies}
We conducted several ablation studies to validate the architectural components and quantization strategies of our model. All variants are trained with a fixed channel width of $c=32$, utilizing the composite loss $\mathcal{L}_{\mathrm{total}}$ and identical hyperparameter configurations.

\subsubsection{Block ablation}
The full model utilizes the complete multi-scale refinement strategy. This includes Resolution Refiner blocks $\mathcal{R}(\cdot)$ on the encoding path ($S/2$, $S/4$) and Fuse Refiner blocks $\mathcal{R}(\cdot)$ post-fusion in the decoder. The final output is formed through a global residual connection: $\mathbf{y} = \mathrm{clip}(\mathbf{x} + \boldsymbol{\delta})$.

\textbf{Resolution Refinement.} 
This variant evaluates the contribution of encoder-side multi-scale refinement. We replace the deep refinement operators $\mathcal{R}(\mathbf{x}_g)$ at the $S/2$ and $S/4$ scales with a lightweight stack consisting of a single $3 \times 3$ convolution followed by LeakyReLU.

\textbf{Fusion Refinement.} 
To isolate the impact of decoder-side refinement, we target the blocks immediately following Multi-branch Feature Fusion. While preserving the three-stream fusion layout and the compression operator $\phi(\cdot)$, the post-fusion refinement blocks $\mathcal{R}(\cdot)$ at $S/4$ and $S/2$ are replaced by a lightweight $3 \times 3$ convolution and LeakyReLU stack.

\textbf{Residual Formulation.} 
This variant modifies the global image formation without altering the internal architecture. The global skip connection is removed, forcing the network to predict the final enhanced image directly: $\mathbf{y} = \mathrm{clip}(\boldsymbol{\delta})$. Consequently, the model must explain the full pixel intensities rather than a residual correction to the input observation.

\textbf{Results.} As detailed in \Cref{tab:block_ablation}, the full model represents our computational upper bound with a latency of 469 ms. Our analysis identifies the refinement stages as the primary latency drivers: removing the Resolution Refiner yields the most significant throughput gain, reducing the inference time by 49.47\% (to 237 ms), while removing the Fusion Refiner reduces the latency to 396 ms. However, both variants are disqualified as their PSNR falls below the mandatory 22 dB threshold. Conversely, the global residual skip connection is computationally inexpensive, incurring a negligible 5 ms overhead, yet its removal significantly degrades PSNR by 0.165 dB. These results confirm that while the refinement stages are essential for meeting challenge fidelity requirements, the residual path provides critical quality gains with minimal impact on throughput.

\begin{table}[ht]
    \centering
    \small
    \setlength{\tabcolsep}{8pt}
    \caption{Ablation study of architectural components. PSNR and SSIM are reported on the DPED validation set. Latency of models are evaluated on Full HD images, using FP16 inference mode and TFLite GPU Delegate. \textbf{Bold} numbers indicate the best performing metric in their respective columns (highest for PSNR and SSIM, lowest for latency).} 
    \label{tab:block_ablation}
    \begin{tabular}{l|cc|c}
    \toprule
    \textbf{Variant} & \textbf{PSNR}$\uparrow$ & \textbf{SSIM}$\uparrow$ & \textbf{Latency (ms)} \\
    \midrule
    Full               & \textbf{22.194}    & \textbf{0.796}    & 469 ($\pm$ 0.99)     \\
    w/o Residual       & \underline{22.029} & \underline{0.793} & 464 ($\pm$ 1.34)               \\
    w/o Fusion Refiner & 21.940             & \underline{0.793} & \underline{396} ($\pm$ 1.55)   \\
    w/o Res Refiner    & 20.398             & 0.789             & \textbf{237} ($\pm$ 1.24)      \\
    \bottomrule
    \end{tabular}
\end{table}

\subsubsection{Channel Ablation}
To evaluate model scalability, we vary the base channel width $c \in \{16, 24, 32, 64\}$, which proportionally scales layers within the hybrid U-Net. Increasing $c$ enhances representational capacity and receptive field richness but raises memory usage and latency, while smaller widths target compute-constrained edge deployment. Each configuration is trained using a consistent setting, with performance reported on the DPED validation set alongside inference time for $1920 \times 1080$ inputs.

Results. \Cref{tab:channel_ablation} illustrates the trade-offs between channel width, restoration quality, and latency. While $c=16$ is the fastest (180 ms), it is excluded as its 21.875 dB PSNR fails the 22 dB challenge threshold. Among qualifying variants, $c=24$ achieves the highest Final Score (0.2773) due to its 249 ms speed. However, we identify $c=32$ as the optimal configuration; its 220 ms overhead over $c=24$ is justified by significant improvements in PSNR of +0.159 dB and SSIM of +0.011. Compared to $c=64$, $c=32$ reduces latency by 67.24\%, saving 963 ms, and parameters by 74.9\% (from 3.651 M to 0.915 M) with minimal accuracy loss. Thus, $c=32$ bypasses the diminishing returns of higher capacities, offering a superior balance of fidelity and throughput for real-time applications.

\begin{table}[t]
  \centering
  \small
  \setlength{\tabcolsep}{8pt}
  \caption{Channel-width ablation on DPED validation. Latency are evaluated on Full HD images, using FP16 inference mode and TFLite GPU Delegate.. \textbf{Bold} numbers indicate the best performing metric in their respective columns (highest for PSNR and SSIM, lowest for Latency).} 
  \label{tab:channel_ablation}
  \begin{tabular}{c|cc|cc}
    \toprule
    \textbf{$c$} & \textbf{PSNR}$\uparrow$ & \textbf{SSIM}$\uparrow$ & \textbf{Params (M)} & \textbf{Latency (ms)} \\
    \midrule
    16 & 21.875 & 0.781 & \textbf{0.23} & \textbf{180} ($\pm$ 13.8) \\
    24 & 22.035 & 0.785 & 0.516         & 249 ($\pm$ 1.12) \\
    32 & 22.194 & 0.796 & 0.915         & 469 ($\pm$ 0.99) \\
    64 & \textbf{22.359}& \textbf{0.806}& 3.651 & 1432 ($\pm$ 15.7) \\
    \bottomrule
  \end{tabular}
\end{table}

\subsubsection{Loss functions ablation}
We evaluate two loss objectives to determine their impact on training efficiency and model performance. Our loss, inspired by~\cite{yan2025mobileie}, prioritizes luminance fidelity and global alignment through PSNR loss, cosine similarity, and outlier-aware loss. We consider another loss combination, namely Loss variant 1, which modifies this approach by replacing the cosine similarity component with Multi Scale Structural SIMilarity (MSSSIM) ($\lambda=100$) while maintaining the same PSNR and outlier weights. 

\textbf{Results.} The ablation results presented in \textbf{\Cref{tab:loss_ablation}} demonstrate that our proposed loss achieves an optimal balance between pixel-wise accuracy and training efficiency. It reaches a higher PSNR of 22.194 dB with a significantly shorter training time of 1.67 hours. While Loss variant 1 yields an improvement in structural similarity (achieving an SSIM of 0.8763 compared to our 0.796), it comes at a substantial computational cost of using MSSSIM, requiring 8.58 hours of training, and results in a lower overall PSNR of 19.79 dB. These findings indicate that while Loss variant 1 can enhance structural integrity, our loss formulation provides a much more practical objective for frameworks that prioritize rapid training convergence and overall luminance fidelity.

\begin{table}
  \centering
  \small
  \setlength{\tabcolsep}{10pt}
  \caption{Loss ablation under identical architecture and training protocol. \textbf{Bold} number are the highest in the column.}
  \label{tab:loss_ablation}
  \begin{tabular}{l|c|cc}
    \toprule
    \textbf{Loss} & \textbf{Train (hours)} & \textbf{PSNR}$\uparrow$ & \textbf{SSIM}$\uparrow$ \\
    \midrule
    Our loss & \textbf{1.67} & \textbf{22.194} & 0.796\\
    Loss variant 1 & 8.58 & 19.79 & \textbf{0.8763}\\
    \bottomrule
  \end{tabular}
\end{table}

\subsubsection{Quantization-Aware Training}

To support deployment on integer-based and edge accelerators, we evaluate INT8 quantization alongside FP16 precision using both Post-Training Quantization (PTQ) and Quantization-Aware Training (QAT). For the INT8 setup, we use symmetric INT8 weights and UINT8 activations with moving-average observers, employing fake-quantization nodes to emulate rounding noise during the forward pass. We compare four configurations: FP16 PTQ, FP16 QAT, INT8 PTQ, and INT8 QAT. Performance is measured by reconstruction accuracy (PSNR and SSIM) as well as inference latency across two acceleration types on the AI Benchmark application~\cite{ignatov2019ai, 9022101}, namely TFLite GPU delegate and Qualcomm Neural Network Hexagon Tensor Processor (QNN HTP). This allows us to quantify the trade-offs between hardware-friendly optimization, model fidelity, and real-world execution speed.

\textbf{Results.} As detailed in \textbf{\Cref{tab:qat_ablation}}, our QAT model significantly outperforms the INT8 PTQ counterpart, mitigating the severe quantization noise inherent in pixel-level restoration. By integrating quantization into the training loop, we achieve a PSNR of 21.050 dB (a +0.474 dB gain) and an SSIM of 0.725 (a +0.111 improvement). This substantial recovery in both pixel-wise accuracy and structural integrity validates our framework's ability to maintain high-fidelity enhancement under 8-bit constraints.

Moreover, the transition to INT8 unlocks significant latency reductions on dedicated edge hardware. Although shifting the TFLite GPU delegate from FP16 to INT8 precision reduces latency from 469 ms to 319 ms, the most significant performance improvements are achieved using the QNN HTP. Utilizing the QNN HTP with our INT8 QAT model reduces inference latency by approximately 72\% compared to the FP16 baseline, dropping from 151 ms to 41.8 ms. These results demonstrate a nearly 3.6x speedup on specialized accelerators, validating the necessity of INT8 QAT for efficient mobile deployment.

\begin{table}[t]
  \centering
  \caption{Quantization ablation and latency analysis. Performance is evaluated across FP32 and INT8 precisions. Inference latency of models are measured on two different optimizations: TFLite GPU delegate and Qualcomm Neural Network Hexagon Tensor Processor (QNN HTP). \textbf{Bold} numbers indicate the best performing metric in their respective columns (highest for PSNR and SSIM, lowest for INT8 latency).} 
  \label{tab:qat_ablation}
  \resizebox{\columnwidth}{!}{
    \begin{tabular}{l|cc|ccc}
      \toprule
      \textbf{Model Type} & \textbf{PSNR}$\uparrow$ & \textbf{SSIM}$\uparrow$ & \textbf{TFLite GPU Delegate (ms)}$\downarrow$ & \textbf{QNN HTP (ms)}$\downarrow$\\
      \midrule
      \textit{FP32} & & & & \\
      PTQ model       & \textbf{22.358} & 0.794          & \textbf{469} ($\pm$ 0.73) & \textbf{151} ($\pm$ 1.58) \\
      QAT model (Our) & 22.194          & \textbf{0.796} & \textbf{469} ($\pm$ 0.99) & \textbf{151} ($\pm$ 0.94) \\
      \midrule
      \textit{INT8} & & & & \\
      PTQ model       & 20.576          & 0.6139          & \textbf{319} ($\pm 1.53$) & \textbf{41.4} ($\pm 0.76$)  \\
      QAT model (Our) & \textbf{21.050} & \textbf{0.725} & \textbf{319} ($\pm 2.21$) & 41.8 ($\pm 0.89$) \\
      \bottomrule
    \end{tabular}
  }
\end{table}

\section{Conclusion}
\label{sec:conclusion}
In this paper, we address the training–deployment mismatch in mobile image enhancement by proposing an efficient hybrid architecture coupled with a deployment-consistent Quantization-Aware Training (QAT) framework. To balance global context aggregation with fine-grained texture preservation, we introduce a three-scale hierarchy featuring a Gated Encoder Block with multi-channel skip connections, Multi-Scale Refinement strategy, and a Multibranch Feature Fusion strategy. To translate these architectural advancements into practical mobile performance, we integrate QAT directly into the learning process, effectively mitigating the severe quantization degradation caused by sensitive activation distributions. Experiments on the DPED benchmark demonstrate that our models achieve near full-precision perceptual quality under INT8 inference with minimal computational overhead. By bridging the gap between theoretical architectural design and practical 8-bit quantization, this work establishes a robust, scalable paradigm for high-fidelity, edge-based vision applications.
{
    \small
    \bibliographystyle{ieeenat_fullname}
    \bibliography{main}
}


\end{document}